\documentclass[12pt,letterpaper]{article}

\usepackage{fullpage}
\usepackage{setspace}
\usepackage{amsmath}
\usepackage{amsfonts}
\usepackage{graphicx}
    \graphicspath{ {./figures/} }

\newcommand{\real}{\mathbb{R}}

\title{Gradient Optimization for Single-State RMDPs}
\author{Keith Badger}

\begin{document}

\maketitle

\begin{abstract} 
As modern problems such as autonomous driving, control of robotic components, and medical diagnostics have become increasingly difficult to solve analytically, data-driven decision-making has seen a large gain in interest. Where there are problems with more dimensions of complexity than can be understood by people, data-driven solutions are a strong option. Many of these methods belong to a subdivision of machine learning known as reinforcement learning. 
Unfortunately, data-driven models often come with uncertainty in how they will perform in the worst of scenarios. Since the solutions are not derived analytically many times, these models will fail unpredictably. In fields such as autonomous driving and medicine, the consequences of these failures could be catastrophic. 

Various methods are being explored to resolve this issue and one of them is known as adversarial learning. It pits two models against each other by having one model optimize its goals as the opposite of the other model's goals. This type of training has the potential to find models which perform reliably in complex and high stakes settings, although it is not certain when this type of training will work. The goal is to gain insight about when these types of models will reach stable solutions.
\end{abstract}
\newpage

\section{Introduction}

\subsection{Motivation}
So let's say we're trying to develop an autonomously driving car as many auto companies are trying to do now. There are many different problems to consider if we're to give this "driver" a low chance of ever causing problems that exceed an acceptable degree of damage. We could try to train this driver based on what human drivers do in certain scenarios. This method could be employed through the use of a neural network. The process would entail feeding information about the environment to the driving agent and regressing its decisions against many different human decisions in the same environment. These methods work well for scenarios that are well covered by the data, but let's say this robot driver were to encounter a scenario that the data did not cover well if at all. For example many drivers stop for bikes or pedestrians going across a crosswalk, but what about people walking their bikes? This scenario happens less frequently and if our data set or environment did not adequately detail the actions associated with seeing a bike-walker going across a crosswalk, the driving agent will still make a confident decision that could be correct, or wrong. If the robot driver decides the bike-walker is nothing more than debris, it will not stop for them. This type of agent would not be very desirable for this problem since across many instances of autonomous vehicles in uncovered situations by the data, we could expect many accidents with irreparable damage. Thus a new model may be more appropriate. 

Instead of trusting the data to account for corner cases in the task you are training for, we should manipulate the data to create the worst possible responses for the neural net we have trained. Once a net is trained, if a noise pattern is applied to whatever input in such a way that it minimizes performance within certain limitations for a given noise additive, then the neural model will typically fail miserably. In image recognition for example, if a neural net is used to categorize the contents of an image, it will perform very well on organic data, which has no manipulations. Once the images are doctored to cause the net to fail, it will select an incorrect category with high confidence when a human would easily identify the image correctly. What's useful about this process is that it identifies weaknesses in the model which would otherwise stand if the net continued to regress against regular labeled data, but by regressing against the doctored cases alongside regular data, the neural net will reach a much more robust solution to the task. This type of training is well understood in the machine learning field, but there is another type of training  that has unique benefits which use these methods and do not have well-understood consequences.

Going back to the self-driving car example, instead of using a data set based on what people experience in everyday driving, let's try and simulate the driving experience on a computer where the consequences for failure are minimal. In this simulation, we could give the robot driver incentives and penalties based on how their performance relates to our goals. Now we are in the realm of reinforcement learning. In this scenario, we don't need data from other drivers, we aren't assuming that we have optimal decisions associated with given scenarios, and there is more choice as to what the agent encounters and how it changes its behavior as a result. Unfortunately given our limited perception of all the potential failures for this robot driver, we cannot manually attend to its goals and expect it to reach a comprehensive model of a good driver. Not only would explicitly setting the goals like where to go and how to get there for the driving agent be tedious and time-consuming but our own bias as to what a driver would need to understand would flaw the overall results of the training. Instead, we could assign another agent the task of setting these training goals with its own goal of causing as much damage and taking as much time as possible, the opposite of our driving agent's goals. In this scenario, if the driving agent had a glaring weakness for driving in certain situations, such as bike walkers crossing the road, the goal-setting agent would exploit them, asking the driving agent to come up to as many crosswalks with bike walkers as it could until the driving agent learned to drive well in these scenarios. Once the driving agent corrected its driving policy the goal-setting agent would lose interest in those road conditions searching for new ones to create problematic results. This process would continue to force the driving agent to patch the holes in its policy, and if it eventually reached a stalemate or a saddle point with the goal-setting agent in terms of performance, then the resulting driving policy would perform very reliably in actual road settings. A major question that comes up, is if these two opposing systems will reach such a point. This depends upon the algorithms behind each of the learner systems and the task they optimize around. For a particular type of algorithm, and problem this convergence is guaranteed~\cite{Chambolle2011,Chambolle2016} and we hope to observe this as we implement these models in simplistic scenarios.

\subsection{Problem Description}

There are several key characteristics of the reinforcement learning design or Markovian Problem setting. The first is that the task and model interface is modeled as a cyclical relationship between an environment and an agent (the algorithm that is being trained). The environment is a set of states that the agent responds to by selecting an action from the set of potential actions. Thus the cycle goes as follows: There is some initial state which is taken by the agent and in return, the agent selects an action. This action is fed into the environment, and the environment transitions the agent into a new state. The agent then selects a new action based on the new state, and the environment changes the state again. This process can continue for some set length or indefinitely. Along with a new state, the environment also returns a reward as a function of the state that the agent was in alongside the action it took in response. This reward is what the agent optimizes around and there are a variety of different methods for doing this.

One of the most successful reinforcement learning methods is ``Policy Gradient'' which optimizes an equation that represents how the reward received from the environment changes as a function of the agent's decision-making algorithm. In many cases, the agent's decision-making is characterized by a set of parameters that correspond to the preferences it has towards certain actions given a set of features in the state. Once an equation that models this relationship is derived, the algorithm's preferences can be optimized around the average reward earned in the environment. This process entails adjusting the agent's parameters proportionally to how they change the reward, which is approximated by the change modeling function.

When two policy gradient algorithms have opposite reward functions for the same environment, they are in an adversarial setup where each will optimize their strategies to directly conflict with the other algorithm's policy. If the two policies reach a balancing point where either one of the agents changing their policy will cause them to perform worse, they are said to be at a \emph{saddle point} or an equilibrium~\cite{Behzadian2021,Ho2021a}. What we're interested in is whether or not reaching this point is guaranteed based on the algorithms and the problem. In particular, we would like to know if certain policy gradient algorithms converge in a variety of complex scenarios which are modified versions of simplistic ones.

\section{Static Problem}
By viewing the relationship between the environment and the agent as a game where the agent maximizes reward, and the environment minimizes it, where the agent is a probability distribution of actions and the environment is a probability distribution of states, with a set reward defined for each state action pair, a simplistic Markovian problem is created. Here the probability of encountering different states is not dependent upon actions taken, nor is the probability of taking different actions dependent on the state. This formulation has the following consequences.

The main objective is to solve the following max-min problem for some reward matrix $A \in \real^{n\times n}$ with opposing player vectors $x$ and $y$ of length $n$:
\[
    \max_{x \in \Delta^n} \, \min_{y \in \Delta^n} \; F(x,y) 
\]
where
\begin{align*}    
    F(x,y) &\;=\; \ x^T A y \\
    \Delta^m &\;=\; \left\{ v \in \real^m \mid \sum_{i=0}^m v_i = 1,\; v \geq 0\right\}
\end{align*}

In order to guarantee that the updated vectors stay in $\Delta^n$, the update formulas take the form:

\begin{align*}
   x^{k+1} &= x^k + \alpha(\bar{x} - x^k) 
\end{align*}

\begin{align*}
   y^{k+1} &= y^k + \alpha(\bar{y} - y^k) 
\end{align*}

with $x^k, y^k \in \Delta^n$ where $\bar{x}$ and $\bar{y}$ are to solutions to a gradient descent subproblem which which optimizes $x \in \Delta^n$ for an objective function $G$, and $\alpha \in [0,1]$ as a stepsize parameter.

A couple of notes for the optimization methods below. By simply taking vectors $\hat{x}$ and $\hat{y}$ as the mean of the iterates $x_k$ and $y_k$ respectively, a stronger solution to original problem is obtained for sufficiently large numbers of iterations.
Since $F(x,y)$ is well defined, computing its gradients is simple:

\begin{eqnarray}
\nabla_x F &=& Ay \\
\nabla_y F &=& A^Tx
\end{eqnarray}

Also a key condition for a constrained minimum $x \in \Delta^m$ of an objective function $G$ is

\begin{equation}
\forall i,j \quad \left(x_i > 0 \implies \frac{\partial G(x_i)}{\partial x_i} \leq \frac{\partial G(x_j)}{\partial x_j}\right)
\end{equation}
Essentially this means that all the non-zero positions of the optimal solution $x$ must have partial derivatives of G equal to the minimum partial derivative of G in the vector. If this were not the case, then a more optimal solution could be obtained by simply decreasing one of the non-zero positions, and increasing a position which has a smaller partial derivative by the same amount. This will be useful for deriving the update formulas of each method.

\section{Conditional Gradient}

The first method, conditional gradient descent also known as Frank-Wolfe, is characterized by the subproblems:
\[
\bar{x} = \min_{x \in \Delta^n} \left[ -\nabla F(x^k)^T(x-x^k)\right]
\]
\[
\bar{y} = \min_{y \in \Delta^n} \left[ \nabla F(y^k)^T(y-y^k)\right]
\]

Let $G$ be the objective function contained in the minimization for x, then an optimal solution must satisfy (3). Since $\frac{\partial G(x_i)}{\partial x_i} = -\nabla F(x^k)$, only coordinates with maximal corresponding gradient terms can be non-zero. There can be multiple solutions for $\bar{x}$, but by simply committing to one term which has a maximal gradient term, a valid solution can consistently be obtained. Formally that is:

\begin{align*}
   \bar{x} &= \left\{ x \in \real^n \mid  \begin{cases} x_i = 1, \text{ if } i = j \\
                                                         x_i = 0, \text{ otherwise } \end{cases} \right\}
\end{align*}
where
\begin{align*}
    j = \arg\;\max_{i=1,...,n} (Ay)_i
\end{align*}
and the same can be done for y with a slight modification. Since $\frac{\partial G(y_i)}{\partial y_i} = \nabla F(y^k)$, only coordinates with minimal corresponding gradient terms can be non-zero. This makes
\begin{align*}
   \bar{y} &= \left\{ y \in \real^n \mid  \begin{cases} y_i = 1, \text{ if } i = j \\
                                                         y_i = 0, \text{ otherwise } \end{cases} \right\}
\end{align*}
where
\begin{align*}
    j = \arg\;\min_{i=1,...,n} (A^Tx)_i
\end{align*}

Below is a figure of
\[F(\hat{x},\hat{y}) = \text{Reward} \]
\[\max_{x \in \Delta^n} F(x,\hat{y}) =  \text{Upperbound} \]
\[\min_{y \in \Delta^n} F(\hat{x},y) = \text{Lowerbound} \]
given a 3x3 grid representing x and y playing rock paper scissors where a win for $x$ is $+1$, a loss is $-1$, and a draw is $0$, plotted over 1000 iterations using Conditional Gradient Descent, setting $\alpha = .01$.

\includegraphics{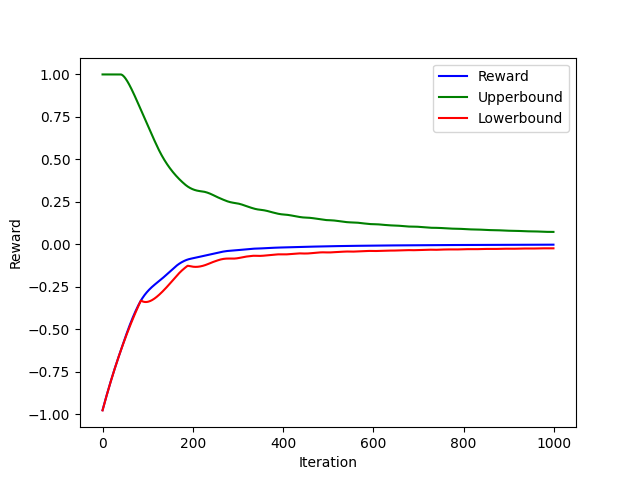}

Here is another figure which represents $x$ and $y$'s path through the 3-dimensional probability simplex for Conditional Gradient Descent where $x$ is green and $y$ is red.

\includegraphics{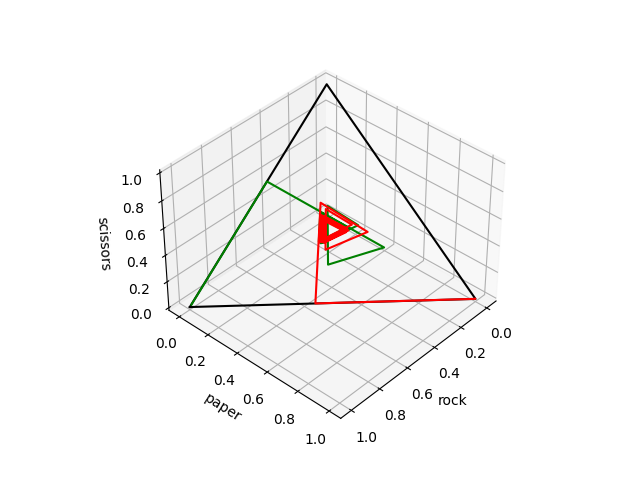}

\section{Euclidean Penalized Gradient}

If instead of just minimizing the dot product of the cost gradient and change in position at each step, we solve
\[
\bar{x} = \min_{x \in \Delta^n} \left[ -\nabla F(x^k)^T(x-x^k) + \beta \cdot \| x - x^k \|_2^2\right]
\]
\[
\bar{y} = \min_{y \in \Delta^n} \left[ \nabla F(y^k)^T(y-y^k) + \beta \cdot \| y - y^k \|_2^2\right]~,
\]
where $\beta \in \real_+$ and $G$ is the objective function minimized by $x$ again, then in order to compute an $\bar{x}$ which satisfies (3), $\frac{\partial G(x_i)}{\partial x_i}$ must be found. That is:
\begin{align*}
    \frac{\partial G(x_i)}{\partial x_i} =& \frac{\partial \left( -\nabla F(x_i^k)^T(x_i-x_i^k) + \beta \cdot \|x_i-x_i^k\|_2^2 \right)}{\partial x_i}
\end{align*}
\begin{align*}
      =& -\nabla F(x_i^k) + 2\beta (x_i - x_i^k)
\end{align*}
\begin{align*}
      =& -\nabla F(x_i^k) - 2\beta x_i^k + 2\beta x_i
\end{align*}
Since the optimization condition above needs all non-zero coordinates to have minimal and equal partial derivatives of $G$, and all values of $x$ must be positive, it can be seen that the terms containing $x_i^k$ will determine the values of $i$ for which $x_i$ will be zero. Specifically, if
\[
d_i = -\nabla F(x_i^k) - 2\beta x_i^k
\]
then
\[
(x_i = 0)  \implies \left( d_j \geq d_i \implies x_j = 0\right)
\]
This is true since if there were an $x_j$ greater than zero for which $d_j$ was greater than $d_i$ where $x_i$ was zero, then the partial derivative of $x_j$ would be larger than $x_i$ and its value would be non-zero, which violates the optimality conditions from before. This idea lends itself to a certain method for computing $\bar{x}$. Since we know $\bar{x}$ will be in the simplex we can solve the non-zero $x_i$ to make all their $\frac{\partial G(x_i)}{\partial x_i}$ equal to some $\lambda$ where $\lambda$ is chosen such that the sum of the $x_i$ is equal to one. If $x_i$ is some nonzero coordinate of the optimal solution $\bar{x}$, then it can be solved in the following fashion:
\begin{align*}
    \lambda &= -\nabla F(x_i^k) - 2\beta x_i^k + 2\beta x_i
\end{align*}
\begin{align*}
    \lambda + 2\beta x_i^k + \nabla F(x_i^k) &= 2\beta x_i
\end{align*}
\begin{align*}
    \frac{\lambda + 2\beta x_i^k + \nabla F(x_i^k)}{2\beta} &= x_i~.
\end{align*}
Now since
\[
\sum_i x_i = 1
\]
it follows that
\begin{align*}
\sum_i \frac{\lambda + 2\beta x_i^k + \nabla F(x_i^k)}{2\beta} &= 1
\end{align*}
\begin{align*}
\sum_i \lambda + 2\beta x_i^k + \nabla F(x_i^k) &= 2\beta
\end{align*}
\begin{align*}
n\lambda &= 2\beta - \sum_i 2\beta x_i^k + \nabla F(x_i^k)
\end{align*}
\begin{align*}
\lambda &= \frac{2\beta - \sum_i 2\beta x_i^k + \nabla F(x_i^k)}{n}~.
\end{align*}
By substituting in $\lambda$
\begin{align*}
x_i &= \frac{\frac{2\beta - \sum_i 2\beta x_i^k + \nabla F(x_i^k)}{n} + 2\beta x_i^k + \nabla F(x_i^k)}{2\beta}
\end{align*}
\begin{align*}
 &= \frac{2\beta - \sum_i [2\beta x_i^k + \nabla F(x_i^k)] + n(2\beta x_i^k + \nabla F(x_i^k))}{n2\beta}
\end{align*}
which is $x_i$ in terms of its previous vector $x^k$, $\beta$, and $n$(the number of nonzero coordinates in $x$). Now since it's not known which coordinates in $\bar{x}$ are zero, an easy way to find out is to start by assuming that all the positions in $\bar{x}$ are non-zero and then solve for each $x_i$. If there are negative values in the solution then at least one of the values in the true solution of $\bar{x}$ is zero. To test if there is exactly one zero in the true solution, all that has to be done is to re-solve $x$ excluding the position corresponding to the greatest $d_i$, since that position is the only position that could be zero with only one $x_i$ equal to zero without violating (3). This process of eliminating the positions of $x$ which correspond to the next largest $d_i$ and re-solving can be continued until none of the positions in the solution are negative. In that case the solution is equal to $\bar{x}$. This process can be used to solve $\bar{y}$ as well if its applied to the y-minimization instead.
\[\]
Below is a figure of
\[F(\hat{x},\hat{y}) = \text{Reward} \]
\[\max_{x \in \Delta^n} F(x,\hat{y}) =  \text{Upperbound} \]
\[\min_{y \in \Delta^n} F(\hat{x},y) = \text{Lowerbound} \]
given a 3x3 grid representing x and y playing rock paper scissors where a win for $x$ is $+1$, a loss is $-1$, and a draw is $0$, plotted over $1000$ iterations using Euclidean Penalized Gradient Descent, setting $\alpha = .01$ and $\beta = .5$.

\includegraphics{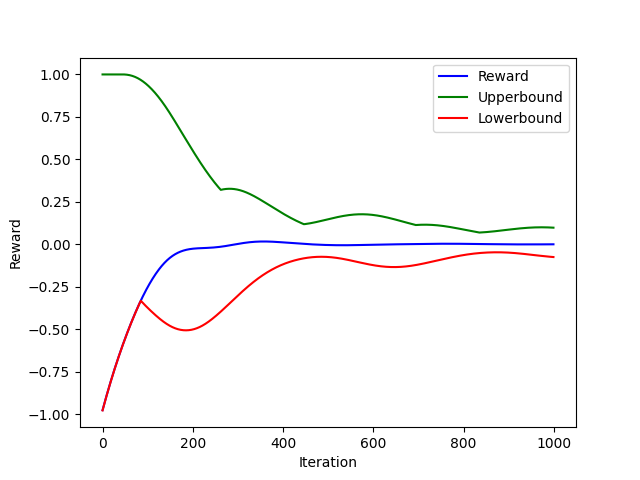}

Here is another figure which represents $x$ and $y$'s path through the 3-dimensional probability simplex for Euclidean Penalized Gradient Descent where $x$ is green and $y$ is red.

\includegraphics{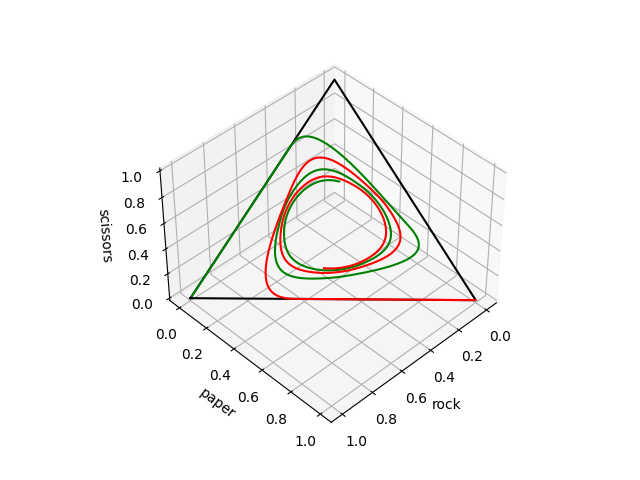}

\section{KL Divergence Penalized Gradient}

If $\beta \cdot \| x - x^k \|_2^2$ was substituted with $\beta \cdot \sum_i x_i \log(\frac{x_i}{x_i^k})$ instead, the new minimization would penalize based on KL divergence of $x$ from $x^k$ instead of the Euclidean distance. Thus the new problems to be solved at each step would be:

\begin{align*}
    \bar{x} = \min_{x \in \Delta^n} \left[ -\nabla F(x^k)^T(x-x^k) + \beta \cdot \sum_i x_i \log \left(\frac{x_i}{x_i^k}\right) \right]~.
\end{align*}
\begin{align*}
    \bar{y} = \min_{y \in \Delta^n} \left[ \nabla F(y^k)^T(y-y^k) + \beta \cdot \sum_i y_i \log \left(\frac{y_i}{y_i^k}\right) \right]~.
\end{align*}

If the same approach for finding the solution $\bar{x}$ is used from earlier, where each partial derivative of the minimization function $G$ is set to some constant $\lambda$ then the values of $x_i$ can be derived as follows:

\begin{align*}
    \lambda &= -\nabla F(x_i^k) + \beta (1 + \log\left(\frac{x_i}{x_i^k}\right))
\end{align*}
\begin{align*}
    \lambda &= -\nabla F(x_i^k) + \beta - \beta\ \log(x_i^k) + \beta\ \log(x_i)
\end{align*}
\begin{align*}
    \lambda + \nabla F(x_i^k) - \beta + \beta\ \log(x_i^k) &= \beta\ \log(x_i)
\end{align*}
\begin{align*}
   \frac{\lambda + \nabla F(x_i^k)}{\beta} - 1 + \log(x_i^k) &= \log(x_i)
\end{align*}
\begin{align*}
   x_i &= \exp\left( \frac{\lambda + \nabla F(x_i^k)}{\beta} - 1 + \log(x_i^k) \right)
\end{align*}
and by solving $\lambda$ such that
\[
\sum_i x_i = 1
\]
\begin{align*}
    \sum_i \exp\left( \frac{\lambda + \nabla F(x_i^k)}{\beta} - 1 + \log(x_i^k) \right) &= 1
\end{align*}
\begin{align*}
    \exp(\frac{\lambda}{\beta})\sum_i \exp\left( \frac{\nabla F(x_i^k)}{\beta} - 1 + \log(x_i^k) \right) &= 1
\end{align*}
\begin{align*}
    \exp(\frac{\lambda}{\beta}) &= \frac{1}{\sum_i \exp\left( \frac{\nabla F(x_i^k)}{\beta} - 1 + \log(x_i^k) \right)}
\end{align*}
\begin{align*}
    \frac{\lambda}{\beta} &= \log\left( \frac{1}{\sum_i \exp\left( \frac{\nabla F(x_i^k)}{\beta} - 1 + \log(x_i^k) \right)} \right)
\end{align*}
\begin{align*}
    \lambda &= \beta\ \log\left( \frac{1}{\sum_i \exp\left( \frac{\nabla F(x_i^k)}{\beta} - 1 + \log(x_i^k) \right)} \right)
\end{align*}

By plugging in the solved $\lambda$
\begin{align*}
   x_i &= \exp\left( \frac{\beta\ \log\left( \frac{1}{\sum_i \exp\left( \frac{\nabla F(x_i^k)}{\beta} - 1 + \log(x_i^k) \right)} \right) + \nabla F(x_i^k)}{\beta} - 1 + \log(x_i^k) \right)
\end{align*}
\begin{align*}
   x_i &= \exp\left( \log\left( \frac{1}{\sum_i \exp\left( \frac{\nabla F(x_i^k)}{\beta} - 1 + \log(x_i^k) \right)} \right)+\frac{\nabla F(x_i^k)}{\beta} - 1 + \log(x_i^k) \right)
\end{align*}
\begin{align*}
   x_i &= \frac{\exp\left( \frac{\nabla F(x_i^k)}{\beta} - 1 + \log(x_i^k) \right)}{\sum_i \exp\left( \frac{\nabla F(x_i^k)}{\beta} - 1 + \log(x_i^k) \right)}
\end{align*}
\begin{align*}
   x_i &= \frac{x_i^k \exp\left( \frac{\nabla F(x_i^k)}{\beta} - 1 \right)}{\sum_i x_i^k \exp\left( \frac{\nabla F(x_i^k)}{\beta} - 1 \right)}
\end{align*}
\begin{align*}
   x_i &= \frac{e^{-1} x_i^k \exp\left( \frac{\nabla F(x_i^k)}{\beta} \right)}{e^{-1}\sum_i x_i^k \exp\left( \frac{\nabla F(x_i^k)}{\beta}\right)}
\end{align*}
\begin{align*}
   x_i &= \frac{x_i^k \exp\left( \frac{\nabla F(x_i^k)}{\beta} \right)}{\sum_i x_i^k \exp\left( \frac{\nabla F(x_i^k)}{\beta}\right)}
\end{align*}

Unlike the Euclidean penalized minimization, by simply setting all the partials of $G$ equal to one another, the values of the resulting $x$ are guaranteed to be positive and sum to one. Therefore, there is no need to check how many zeros will be in the true solution, since the first $x$ yielded from the above formula will be in the n-dimensional simplex and meet (3), and thus equal $\bar{x}$. The same procedure can be used for solving $\bar{y}$ by replacing the components from the x-minimization with the components of the y-minimization.
\[\]
Below is a figure of
\[F(\hat{x},\hat{y}) = \text{Reward} \]
\[\max_{x \in \Delta^n} F(x,\hat{y}) =  \text{Upperbound} \]
\[\min_{y \in \Delta^n} F(\hat{x},y) = \text{Lowerbound} \]
given a 3x3 grid representing x and y playing rock paper scissors where a win for $x$ is $+1$, a loss is $-1$, and a draw is $0$, plotted over $1000$ iterations using KL Divergence Penalized Gradient Descent, setting $\alpha = .01$ and $\beta = .25$.

\includegraphics{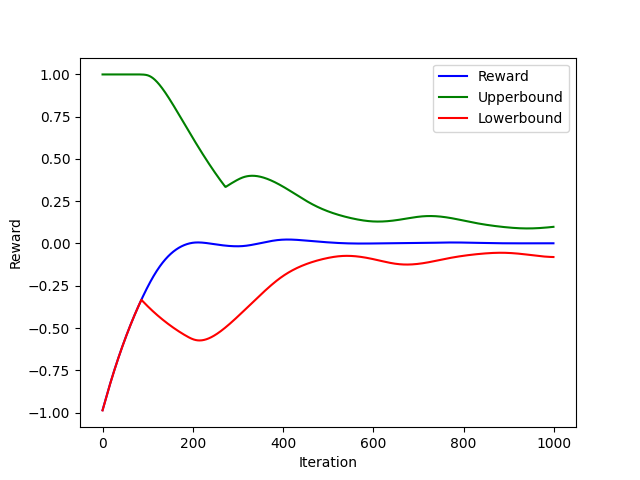}

Here is another figure which represents $x$ and $y$'s path through the 3-dimensional probability simplex for KL Divergence Penalized Gradient Descent where $x$ is green and $y$ is red.

\includegraphics{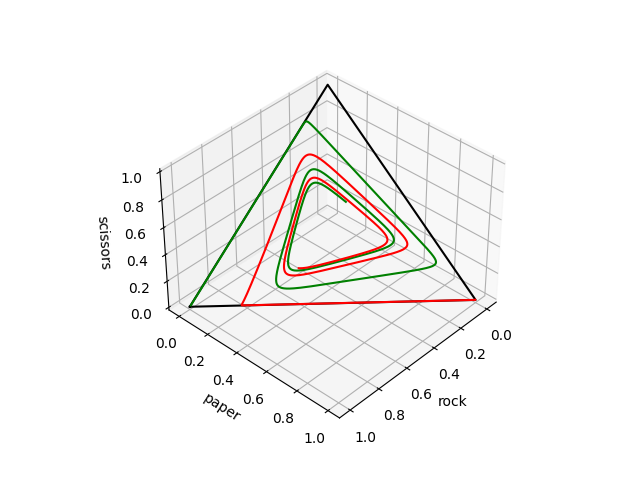}

\section{Future Work}
Although each of these methods solves the max-min of reward for static distributions of actions and states, this scenario does not adequately encapsulate most Markovian problems. Usually the probability of encountering different states is dependent on the actions taken, and the actions taken should depend on the state the agent is in. There are more complex models known as Distributionally Robust Markov Decision Processes which account for these considerations. Instead of assuming a static distribution of states, a probability distribution of next states is assigned to each state-action pair, and instead of the adversarial environment selecting a distribution of states, it selects a distribution of transition rates across the different state action pairs. The agent also has separate probabilities of selecting different actions in different states. Each of these additions to the problem adds another layer of complexity onto the previously solved static problem, but the iterative methods for solving these problems remain similar.

In order to understand the consequences for making the model more applicable, we would need to spend time looking at the current literature describing these systems, implement the algorithms to solve them, and compare their results against our own theoretical results and see how much they differ. If these general purpose algorithms could have better understood consequences, it would have large implications for the many fields that depend on these models.

\bibliographystyle{plain}
\bibliography{Adversarial_Gradient}
\end{document}